\documentclass{article}

\usepackage[preprint]{neurips_2024}

\usepackage[utf8]{inputenc} 
\usepackage[T1]{fontenc}    
\usepackage{hyperref}       
\usepackage{url}            
\usepackage{booktabs}       
\usepackage{amsfonts}       
\usepackage{nicefrac}       
\usepackage{microtype}      
\usepackage{xcolor}         

\usepackage[pdftex]{graphicx}

\title{Adaptive Alignment: Dynamic Preference Adjustments via Multi-Objective Reinforcement Learning for Pluralistic AI}

\author{%
  Hadassah~Harland\\
  Deakin University\\
  Geelong, Australia \\
  \texttt{h.harland@research.deakin.edu.au} \\
  \And
  Richard Dazeley \\
  Deakin University\\
  Geelong, Australia \\
  \texttt{richard.dazeley@deakin.edu.au} \\
  \And
  Peter Vamplew \\
  Federation University \\
  Ballarat, Australia \\
  \texttt{p.vamplew@federation.edu.au} \\
  \And
  Hashini Senaratne\\
  CSIRO \\
  Pullenvale, Australia \\
  \texttt{Hashini.Senaratne@data61.csiro.au} \\
  \And
  Bahareh Nakisa \\
  Deakin University\\
  Geelong, Australia \\
  \texttt{bahar.nakisa@deakin.edu.au} \\
  \And
  Francisco Cruz \\
  University of New South Wales \\
  Sydney, Australia \\
  \texttt{f.cruz@unsw.edu.au} \\
}

\begin{document}

\maketitle

\begin{abstract}
Emerging research in Pluralistic Artificial Intelligence (AI) alignment seeks to address how intelligent systems can be designed and deployed in accordance with diverse human needs and values.
We contribute to this pursuit with a dynamic approach for aligning AI with diverse and shifting user preferences through Multi-Objective Reinforcement Learning (MORL), via post-learning policy selection adjustment. 
In this paper, we introduce the proposed framework for this approach, outline its anticipated advantages and assumptions, and discuss technical details about the implementation. 
We also examine the broader implications of adopting a retroactive alignment approach through the sociotechnical systems perspective. 
\end{abstract}

\section{Introduction}

\textit{Pluralistic} alignment has recently emerged as an area of growing interest within Artificial Intelligence (AI) research~\citep{Sorensen2023, Sorensen2024}. The term unifies a set of ideas from a developing theme regarding the diverse, multifaceted, and evolving nature of human values and the corresponding challenges this presents to human-aligned AI~\citep{Jain2024, Vamplew2018}. 
Given the pluralistic nature of human needs and preferences, human-aligned AI systems must be designed with the capability to autonomously and independently adapt to fit individual users, use cases, and contexts. 

Multi-objective reinforcement learning (MORL) is a powerful AI technique for autonomous sequential decision-making tasks involving multiple, often conflicting, objectives~\citep{Hayes2022}. Through multi-policy techniques, MORL algorithms can learn a spectrum of potential solutions in parallel, each optimised for different objective trade-offs, thereby providing a diverse set of policies that can be dynamically selected at runtime. This adaptability and capacity for balancing competing objectives presents MORL as a promising platform for pluralistic alignment research. 

This paper presents a MORL-based approach to pluralistic AI via \textit{adaptive alignment}, using retroactive policy selection adjustments to continuously realign to the user's preferences. First, we briefly review AI alignment research, exploring key challenges and highlighting the need for a multi-objective approach. In section~\ref{approach}, we present an adaptive alignment framework, which is described in three stages: \textit{learning}, \textit{selection}, and \textit{execution and review}. Then, we discuss technical considerations related to the implementation in Section~\ref{technical}. We conclude by examining the implications of retroactive adjustment in AI alignment, emphasising that the need for post-interaction realignment is unavoidable and the importance of active transparency when interacting with human users.

\section{Challenges in AI alignment} \label{challenges}

Research in AI alignment has grown increasingly critical as AI systems continue gaining ability and prevalence~\citep{Ji2023, Taylor2020}. \citet{Ji2023} describe these efforts according to two main streams; \textit{forward alignment} considers how to design new systems that meet these demands, whereas \textit{backward alignment} looks to ensure alignment of existing systems through regulation, governance, and assurance. 

Reinforcement learning (RL)-based approaches feature prominently in forward alignment~\citep{Ji2023}, leveraging the premise of learning an optimal policy by seeking to maximise an expected cumulative reward~\citep{Sutton2018}.
In particular, \textit{Reinforcement Learning from Human Feedback} is a popular approach~\citep{Ouyang2022}, where the reward function is derived from human preferences. 
These models can be criticised as resource intensive, both computationally and in requiring manually labelled data~\citep{Cao2024, Casper2023}, leading to the emergence of alternative approaches for automating alignment. 
For example, the \textit{Constitutional AI}~\citep{Bai2022} and \textit{Reinforcement Learning from AI Feedback}~\citep{LeeH2023} algorithms replace the human in the training loop with another AI model to enable self-improvement with fewer instances of human feedback. 

However, these approaches can oversimplify the alignment problem by not accounting for the pluralistic nature of human values~\citep{Sorensen2024}. 
The needs and values of different people may vary broadly; even for a single individual across differing contexts, the exact requirements cannot be universally defined~\citep{Gabriel2020, Mishra2023}. 
By resolving to a single, static solution, these algorithms leave no space to accommodate the natural variability in values and preferences between users and contexts. 
Furthermore, without the ability to adapt, these solutions may become outdated as preferences change over time. 

\textit{Adaptive alignment} may help to address this limitation, enabling pluralistic system expression to represent diverse human values and perspectives. 
Thus far, interactive machine learning approaches such as \textit{In-context}~\citep{Dong2022} and \textit{Active}~\citep{Taylor2021} Learning have largely focused on task generalisation. However, some RL-based approaches have recently emerged, enabling adaptive value alignment by representing the task as a Multi-Objective Markov Decision Process (MOMDP) and employing MORL techniques~\citep{Harland2023, Peschl2021, Rame2023, Yang2024}. 

The need for multi-objective approaches for human-alignment in RL is well established~\citep{Casper2023, Mannion2021, Vamplew2018}, as \textit{a priori} scalarisation of objectives does not allow for the necessary exploration, visibility, or flexibility of the solution to support alignment~\citep{Hayes2022}. Conversely, by representing human values as distinct objectives, the agent can separately evaluate and balance competing priorities, enabling it to explore a diverse range of potential solutions~\citep{Vamplew2018}. For example, whether to prioritise cleaning or avoid disruptions~\citep{Harland2023, Peschl2021}, or how to balance humour, helpfulness, and harmlessness in a chatbot response~\citep{Yang2024}. 
Of particular relevance to pluralistic alignment, MORL enables multi-policy learning, such that the specific policy to be executed can be selected \textit{a posteriori} to the learning process~\citep{Hayes2022}. 

Yet, a major challenge remains; how may a suitable policy be selected given potentially unknown and dynamic user preferences? 
\citet{Hayes2022} propose the \textit{review and adjust} scenario to address how a MORL system may adapt to dynamic user preferences, describing a process of retroactive updates via manual user selection.
We propose an extension to this scenario with an approach that circumvents the manual selection process to dynamically adapt to diverse user preferences.

 \section{An adaptive alignment framework} \label{approach}
\begin{figure}[h]
  \centering
  \includegraphics[scale=0.7]{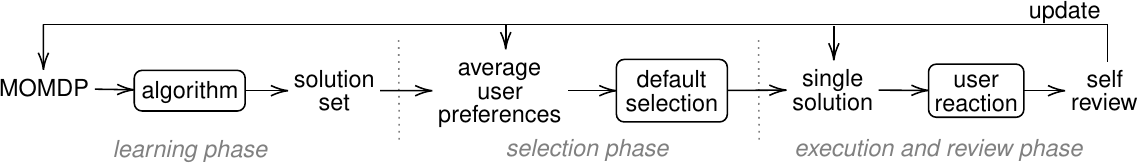}
  \caption{The \textit{self-review} process leverages indirect feedback to adjust to the users' preferences. } 
  \label{self_review}
\end{figure}

In this section, we introduce a framework for pluralistic AI through adaptive alignment in MORL, modelled after the \textit{review and adjust} process (Section~\ref{challenges}). The proposed agent adapts to the user's preferences through a \textit{self-review} process, using context and informal signals to minimise the need for direct and specific feedback from the user (Figure~\ref{self_review}). 
Two key features are distinct: an initial default policy is chosen in the selection phase, and the role of reviewer is shifted from the user to the agent. %

The basis of this framework is a trained, multi-objective, multi-policy RL algorithm that has learned a set of solutions representing the scope of possible human preferences across multiple different values. The algorithm represents these values as distinct objectives, and each solution describes an optimal policy for a particular set of preferences over these objectives. 

The initial policy is selected according to the predicted best fit for the user. For an unknown user, this may either be a universal default selection, or can be initially personalised according to what information is available; for example, the system might prioritise brevity over detail when responding to a voice query. For a familiar user, the choice of policy is informed by previous interactions. 

With each execution, the agent observes the user's reaction (e.g., facial expression, nonverbal audio) and performs a self-review. 
The process draws on information collected about the interaction and relevant contextual factors to identify any misalignment between the current policy and the user's preferences. 
The agent then selects a new policy to dynamically adjust its behaviour accordingly. 

We anticipate some of the advantages to this approach to be the following: 

\textbf{Feedback efficiency and focus:} 
The approach alleviates the need for explicit feedback by using the user's reaction as a signal, which should be less burdensome on the user and minimise the influence of response bias. Furthermore, the brevity of the feedback signal provides a narrow focus, which should inherently reduce the dimensionality of the feedback according to its importance to the user.
    
\textbf{Aligning with multiple users:} 
This framework has been designed with multiple users in mind. If the current user changes, the previous user's preferred policy could be stored when the new user's profile is created or loaded, so the system can retain what it has learned about each previous user while operating according to the new user's preferences.    

\textbf{Continuous evolution:} 
The repeating self-review process provides a continuous feedback loop that should enable the system to accommodate new preferences as they arise and so maintain alignment with the users' evolving needs. By also updating the average users' preferences, the system should also be able to improve the initial select for new users and evolve at a broader social level.

\section{Techniques for adaptive alignment} \label{technical}
The framework described in the previous section relies on two key assumptions: 1) a suitable model can be developed to accurately detect and attribute misalignment in the system based on the information available to the agent at execution, and 2) given the output from this model, the system can perform an update by selecting an alternative policy that is better aligned with the user's preferences.
In this section, we discuss specific techniques to address these assumptions.

\subsection{Interpreting user reactions as feedback}
To satisfy the first assumption, we require an interpretation model that enables the user's reaction to act as a feedback signal. That is, we want to find a model $M$ to transform a reaction signal $\zeta$ into an update $\vec{\Delta\Xi}$ to the user's preferences $\vec{\Xi}$. 
This model should incorporate information about the interaction ($\theta$) so that the signal can be interpreted in context ($M \rightarrow M(\theta)$). Constituent factors should include the outcome of the execution~\citep{MacGlashan2017}, as well as the usual distribution of user preferences for the given use case, and the history of any prior interactions with this user. 

One possible approach could be to define $\vec{\Delta}$ explicitly using a loss measure derived from individually \textit{idealised} reward values $R^{ideal}_i$ for each objective $i$. For this example, we assume the reaction signal $\zeta \in \mathcal{N}(\mu,\sigma^2)$ to be a normally distributed scalar, but other forms are possible~\citep{Jeon2020}. Equation \ref{explicit_model} provides an example, given $\hat{\zeta} \in \mathcal{N}(0,1)$ transformed via Bayesian estimation, scaling factor $\alpha_i$, and activation threshold $\tau_i$.

\begin{equation}
    \Delta_i = \alpha_i \hat{\zeta} (R^{observed}_{i} - R^{ideal}_i) - \tau_i \quad \forall i
    \label{explicit_model}
\end{equation}

An alternative approach could be to employ a RL algorithm by representing the task as a contextual bandit problem~\citep{Bouneffouf2020}. The model would use the context $\theta$ and a reward signal derived from the transformed reaction $\hat{\zeta}$.  
Similar models have previously been used for simulating cognitive empathy in human-robot interaction~\citep{Bagheri2021}.

\subsection{Solution updates via post-learning policy selection adjustment}
To satisfy the second assumption, we require a process to select a new policy $\pi' \in \Pi$ that best aligns with the user's preferences $\Xi$ from the set of known Pareto-optimal policies $\Pi$~\citep{Hayes2022}. 
Possible approaches for identifying $\pi'$ depend on how $\Pi$ is represented. 
Pareto-based methods, such as Pareto Q-Learning~\citep{Moffaert2014}, store learned policies as vector returns. The format eases direct policy comparison at the cost of high computational complexity to reproduce a policy based on its expected return~\citep{Felten2024}. Conversely, approximate methods, such as conditioned networks~\citep{Abels2019}, may learn a parametric policy representation $\pi(\phi)$, or use interpolation to compute a mixture policy using a weighted combination of learned policies~\citep{Rame2023, Yang2024}.

If each policy in $\Pi$ can be mapped directly to a return vector, it is possible to calculate an ordering over the policies using a utility function $u$ derived from $\Xi$. 
The definition of$u$ could be as simple as applying $\vec{\Xi}$ directly as weights for linear scalarisation~\citep{Hayes2022}, or may incorporate non-linear features such as thresholds and lexicographical ordering~\citep{Harland2023}. 

If a direct ordering over the policies is not feasible, it might be more suitable to employ a steering-based approach~\citep{Vamplew2017b}. Instead of selecting a completely new policy, steering enables stepwise updates to the policy selection, by moving along the Pareto front to the next closest policy or mixture of policies in the direction of the update $\vec{\Delta\Xi}$. This approach benefits from a smaller search space and may appear more stable to the user due to the progressive updates, although it may be slower to implement large-scale changes.

The agent adapts its behaviour by executing the updated policy selection $\pi'$. The update itself is strictly not a learning process, as the underlying policies are fixed.  
However, this update process might also help inform aspects of earlier phases (Figure~\ref{self_review}): contributing additional data towards the average user preferences to continuously adapt the default selection, and providing an indication of possible objectives not captured in the MOMDP.

\section{Implications of a retroactive approach} \label{implications}
The adaptive alignment framework we proposed in this paper follows a retroactive approach to pluralistic AI, with some accompanying implications. We consider these implications through the sociotechnical systems perspective;  in matters related to human users, AI algorithms are inseparable from the sociotechnical systems within which they are embedded~\citep{Kudina2024}. 

\textbf{Technical challenges for safety:} 
As noted by \citet{Ji2023}, algorithms that learn through human feedback may be particularly susceptible to risks of reward hacking and scalable oversight~\citep{Amodei2016}. This could be further aggravated by a self-supervisory method such as we have described that may allow potential issues to be obscured. To minimise this risk, it may be beneficial to incorporate backward alignment features such as explanations to provide transparency on how update decisions are made.  

\textbf{Inevitability of misalignment:} 
Prior to the first interaction with a given user, it is not possible to know that user's preferences perfectly. This fact goes beyond any question of feasibility; even if you were to assume the most advanced superintelligence conceivable, it is not philosophically impossible to predict a user's preferences with absolute certainty. Thus, there will always be a need for AI systems to perform retroactive corrections to realign with the evolving needs of the user. The framework proposed herein is an example of one such system, but it does not need to be used in isolation. Rather, this approach should be combined with suitable and effective predictive alignment techniques to minimise the use of adaptive alignment to only where it is necessary. 

\textbf{The need for explanations and repair:} 
The retroactive approach relies on an information signal provided by the user to identify the discrepancy between the current settings and the user's true preferences. This necessitates that misaligned behaviour has already occurred, and corrective action alone may be insufficient to address any harms incurred. Furthermore, users may also adapt their own behaviour throughout the process of interacting with a system as they develop an understanding of how the system behaves. The system may adapt after initially misaligned behaviour, but any subsequent solution will be one step behind, calibrated according to the context of the previous interaction. Thus, when employing a retroactive approach for alignment, it may be appropriate to incorporate both reparations and explanations to support the needs of the user at the sociotechnical scale.

\begin{ack}

This research was supported by an Australian Government Research Training Program (RTP) and a Commonwealth Scientific and Industrial Research Organisation (CSIRO) Top-Up Scholarship.

\end{ack}

{
\small

\bibliographystyle{plainnat}
\bibliography{references} 

\begin{thebibliography}{32}
\providecommand{\natexlab}[1]{#1}
\providecommand{\url}[1]{\texttt{#1}}
\expandafter\ifx\csname urlstyle\endcsname\relax
  \providecommand{\doi}[1]{doi: #1}\else
  \providecommand{\doi}{doi: \begingroup \urlstyle{rm}\Url}\fi

\bibitem[Abels et~al.(2019)Abels, Roijers, Lenaerts, Now{\'{e}}, and Steckelmacher]{Abels2019}
Axel Abels, Diederik~M Roijers, Tom Lenaerts, Ann Now{\'{e}}, and Denis Steckelmacher.
\newblock {Dynamic Weights in Multi-Objective Deep Reinforcement Learning}.
\newblock In \emph{Proceedings of the 36th International Conference on Machine Learning}, 2019.

\bibitem[Amodei et~al.(2016)Amodei, Olah, Steinhardt, Christiano, Schulman, and Man{\'{e}}]{Amodei2016}
Dario Amodei, Chris Olah, Jacob Steinhardt, Paul Christiano, John Schulman, and Dan Man{\'{e}}.
\newblock {Concrete Problems in AI Safety}.
\newblock \emph{ArXiv preprint arXiv:1606.06565}, 2016.

\bibitem[Bagheri et~al.(2021)Bagheri, Roesler, Cao, and Vanderborght]{Bagheri2021}
Elahe Bagheri, Oliver Roesler, Hoang~Long Cao, and Bram Vanderborght.
\newblock {A Reinforcement Learning Based Cognitive Empathy Framework for Social Robots}.
\newblock \emph{International Journal of Social Robotics}, 13\penalty0 (5):\penalty0 1079--1093, 8 2021.
\newblock ISSN 18754805.
\newblock \doi{10.1007/s12369-020-00683-4}.

\bibitem[Bai et~al.(2022)Bai, Kadavath, Kundu, Askell, Kernion, Jones, Chen, Goldie, Mirhoseini, McKinnon, Chen, Olsson, Olah, Hernandez, Drain, Ganguli, Li, Tran-Johnson, Perez, Kerr, Mueller, Ladish, Landau, Ndousse, Lukosuite, Lovitt, Sellitto, Elhage, Schiefer, Mercado, DasSarma, Lasenby, Larson, Ringer, Johnston, Kravec, Showk, Fort, Lanham, Telleen-Lawton, Conerly, Henighan, Hume, Bowman, Hatfield-Dodds, Mann, Amodei, Joseph, McCandlish, Brown, and Kaplan]{Bai2022}
Yuntao Bai, Saurav Kadavath, Sandipan Kundu, Amanda Askell, Jackson Kernion, Andy Jones, Anna Chen, Anna Goldie, Azalia Mirhoseini, Cameron McKinnon, Carol Chen, Catherine Olsson, Christopher Olah, Danny Hernandez, Dawn Drain, Deep Ganguli, Dustin Li, Eli Tran-Johnson, Ethan Perez, Jamie Kerr, Jared Mueller, Jeffrey Ladish, Joshua Landau, Kamal Ndousse, Kamile Lukosuite, Liane Lovitt, Michael Sellitto, Nelson Elhage, Nicholas Schiefer, Noemi Mercado, Nova DasSarma, Robert Lasenby, Robin Larson, Sam Ringer, Scott Johnston, Shauna Kravec, Sheer~El Showk, Stanislav Fort, Tamera Lanham, Timothy Telleen-Lawton, Tom Conerly, Tom Henighan, Tristan Hume, Samuel~R. Bowman, Zac Hatfield-Dodds, Ben Mann, Dario Amodei, Nicholas Joseph, Sam McCandlish, Tom Brown, and Jared Kaplan.
\newblock {Constitutional AI: Harmlessness from AI Feedback}.
\newblock \emph{ArXiv preprint arXiv:2212.08073}, 2022.

\bibitem[Bouneffouf et~al.(2020)Bouneffouf, Rish, and Aggarwal]{Bouneffouf2020}
Djallel Bouneffouf, Irina Rish, and Charu Aggarwal.
\newblock {Survey on Applications of Multi-Armed and Contextual Bandits}.
\newblock In \emph{2020 IEEE Congress on Evolutionary Computation (CEC)}, pages 1--8. IEEE, 7 2020.
\newblock ISBN 978-1-7281-6929-3.
\newblock \doi{10.1109/CEC48606.2020.9185782}.

\bibitem[Cao et~al.(2024)Cao, Lu, Lu, Chen, Ren, Xiang, Liu, Lu, He, Han, Sun, Lin, and Yu]{Cao2024}
Boxi Cao, Keming Lu, Xinyu Lu, Jiawei Chen, Mengjie Ren, Hao Xiang, Peilin Liu, Yaojie Lu, Ben He, Xianpei Han, Le~Sun, Hongyu Lin, and Bowen Yu.
\newblock {Towards Scalable Automated Alignment of LLMs: A Survey}.
\newblock \emph{ArXiv preprint arXiv:2406.01252}, 2024.

\bibitem[Casper et~al.(2023)Casper, Davies, Shi, Gilbert, Tech, Scheurer, Research, Rando, Zurich, Freedman, Korbak, Lindner, Freire, Wang, Marks, Carroll, Peng, Christoffersen, Slocum, Csail, Anwar, Siththaranjan, Nadeau, Michaud, Pfau, Krasheninnikov, Chen, Langosco, Bıyık, Krueger, Sadigh, and Hadfield-Menell]{Casper2023}
Stephen Casper, Xander Davies, Claudia Shi, Thomas~Krendl Gilbert, Cornell Tech, Jérémy Scheurer, Apollo Research, Javier Rando, Eth Zurich, Rachel Freedman, Berkeley~Tomasz Korbak, David Lindner, Pedro Freire, Tony Wang, Samuel Marks, Micah Carroll, Andi Peng, Phillip Christoffersen, Stewart Slocum, Mit Csail, Usman Anwar, Anand Siththaranjan, Max Nadeau, Eric~J Michaud, Jacob Pfau, Dmitrii Krasheninnikov, Xin Chen, Lauro Langosco, Erdem Bıyık, David Krueger, Dorsa Sadigh, and Dylan Hadfield-Menell.
\newblock {Open Problems and Fundamental Limitations of Reinforcement Learning from Human Feedback}.
\newblock \emph{Transactions on Machine Learning Research}, 2023.
\newblock ISSN 2835-8856.
\newblock URL \url{https://openreview.net/forum?id=bx24KpJ4Eb}.

\bibitem[Dong et~al.(2022)Dong, Li, Dai, Zheng, Ma, Li, Xia, Xu, Wu, Chang, Sun, and Sui]{Dong2022}
Qingxiu Dong, Lei Li, Damai Dai, Ce~Zheng, Jingyuan Ma, Rui Li, Heming Xia, Jingjing Xu, Zhiyong Wu, Baobao Chang, Xu~Sun, and Zhifang Sui.
\newblock {A Survey on In-context Learning}.
\newblock \emph{ArXiv preprint arXiv:2301.00234}, 2022.

\bibitem[Felten(2024)]{Felten2024}
Florian Felten.
\newblock \emph{{Multi-Objective Reinforcement Learning}}.
\newblock PhD thesis, University of Luxembourg, 7 2024.

\bibitem[Gabriel(2020)]{Gabriel2020}
Iason Gabriel.
\newblock {Artificial Intelligence, Values, and Alignment}.
\newblock \emph{Minds and Machines}, 30\penalty0 (3):\penalty0 411--437, 9 2020.
\newblock ISSN 15728641.
\newblock \doi{10.1007/s11023-020-09539-2}.

\bibitem[Harland et~al.(2023)Harland, Dazeley, Nakisa, Cruz, and Vamplew]{Harland2023}
Hadassah Harland, Richard Dazeley, Bahareh Nakisa, Francisco Cruz, and Peter Vamplew.
\newblock {AI apology: interactive multi-objective reinforcement learning for human-aligned AI}.
\newblock \emph{Neural Computing and Applications}, 35\penalty0 (23):\penalty0 16917--16930, 8 2023.
\newblock ISSN 0941-0643.
\newblock \doi{10.1007/s00521-023-08586-x}.
\newblock URL \url{https://link.springer.com/10.1007/s00521-023-08586-x}.

\bibitem[Hayes et~al.(2021)Hayes, R{\u{a}}dulescu, Bargiacchi, K{\"{a}}llstr{\"{o}}m, Macfarlane, Reymond, Verstraeten, Zintgraf, Dazeley, Heintz, Howley, Irissappane, Mannion, Now{\'{e}}, Ramos, Restelli, Vamplew, and Roijers]{Hayes2022}
Conor~F. Hayes, Roxana R{\u{a}}dulescu, Eugenio Bargiacchi, Johan K{\"{a}}llstr{\"{o}}m, Matthew Macfarlane, Mathieu Reymond, Timothy Verstraeten, Luisa~M. Zintgraf, Richard Dazeley, Fredrik Heintz, Enda Howley, Athirai~A. Irissappane, Patrick Mannion, Ann Now{\'{e}}, Gabriel Ramos, Marcello Restelli, Peter Vamplew, and Diederik~M. Roijers.
\newblock {A Practical Guide to Multi-Objective Reinforcement Learning and Planning}.
\newblock \emph{Autonomous Agents and Multi-Agent Systems}, 36\penalty0 (1), 3 2021.
\newblock \doi{10.1007/s10458-022-09552-y}.
\newblock URL \url{http://arxiv.org/abs/2103.09568 http://dx.doi.org/10.1007/s10458-022-09552-y}.

\bibitem[Jain et~al.(2024)Jain, Suriyakumar, Creel, and Wilson]{Jain2024}
Shomik Jain, Vinith Suriyakumar, Kathleen Creel, and Ashia Wilson.
\newblock {Algorithmic Pluralism: A Structural Approach To Equal Opportunity}.
\newblock In \emph{2024 ACM Conference on Fairness, Accountability, and Transparency, FAccT 2024}, pages 197--206. Association for Computing Machinery, Inc, 6 2024.
\newblock ISBN 9798400704505.
\newblock \doi{10.1145/3630106.3658899}.

\bibitem[Jeon et~al.(2020)Jeon, Milli, and Dragan]{Jeon2020}
Hong~Jun Jeon, Smitha Milli, and Anca Dragan.
\newblock {Reward-rational (implicit) choice: A unifying formalism for reward learning}.
\newblock In \emph{Thirty-Fourth Conference on Neural Information Processing Systems}, 2020.

\bibitem[Ji et~al.(2023)Ji, Qiu, Chen, Zhang, Lou, Wang, Duan, He, Zhou, Zhang, Zeng, Ng, Dai, Pan, O'Gara, Lei, Xu, Tse, Fu, McAleer, Yang, Wang, Zhu, Guo, and Gao]{Ji2023}
Jiaming Ji, Tianyi Qiu, Boyuan Chen, Borong Zhang, Hantao Lou, Kaile Wang, Yawen Duan, Zhonghao He, Jiayi Zhou, Zhaowei Zhang, Fanzhi Zeng, Kwan~Yee Ng, Juntao Dai, Xuehai Pan, Aidan O'Gara, Yingshan Lei, Hua Xu, Brian Tse, Jie Fu, Stephen McAleer, Yaodong Yang, Yizhou Wang, Song-Chun Zhu, Yike Guo, and Wen Gao.
\newblock {AI Alignment: A Comprehensive Survey}.
\newblock \emph{ArXiv preprint arXiv:2310.19852}, 2023.

\bibitem[Kudina and van~de Poel(2024)]{Kudina2024}
Olya Kudina and Ibo van~de Poel.
\newblock {A sociotechnical system perspective on AI}.
\newblock \emph{Minds and Machines}, 34\penalty0 (3):\penalty0 21, 6 2024.
\newblock ISSN 1572-8641.
\newblock \doi{10.1007/s11023-024-09680-2}.

\bibitem[Lee et~al.(2023)Lee, Phatale, Mansoor, Mesnard, Ferret, Lu, Bishop, Hall, Carbune, Rastogi, and Prakash]{LeeH2023}
Harrison Lee, Samrat Phatale, Hassan Mansoor, Thomas Mesnard, Johan Ferret, Kellie Lu, Colton Bishop, Ethan Hall, Victor Carbune, Abhinav Rastogi, and Sushant Prakash.
\newblock {RLAIF: Scaling Reinforcement Learning from Human Feedback with AI Feedback}.
\newblock \emph{ArXiv preprint arXiv:2309.00267}, 2023.

\bibitem[MacGlashan et~al.(2017)MacGlashan, Ho, Loftin, Peng, Wang, Roberts, Taylor, and Littman]{MacGlashan2017}
James MacGlashan, Mark~K Ho, Robert Loftin, Bei Peng, Guan Wang, David~L Roberts, Matthew~E Taylor, and Michael~L Littman.
\newblock {Interactive Learning from Policy-Dependent Human Feedback}.
\newblock In Doina Precup and Yee~Whye Teh, editors, \emph{Proceedings of the 34th International Conference on Machine Learning}, volume~70 of \emph{Proceedings of Machine Learning Research}, pages 2285--2294. PMLR, 9 2017.
\newblock URL \url{https://proceedings.mlr.press/v70/macglashan17a.html}.

\bibitem[Mannion et~al.(2021)Mannion, Heintz, Karimpanal, and Vamplew]{Mannion2021}
Patrick Mannion, Fredrik Heintz, Thommen~George Karimpanal, and Peter Vamplew.
\newblock {Multi-Objective Decision Making for Trustworthy AI}.
\newblock In \emph{Proceedings of the Multi-Objective Decision Making (MODeM) Workshop}, 2021.

\bibitem[Mishra(2023)]{Mishra2023}
Abhilash Mishra.
\newblock {AI Alignment and Social Choice: Fundamental Limitations and Policy Implications}.
\newblock \emph{SSRN Electronic Journal}, 2023.
\newblock \doi{10.2139/ssrn.4605509}.

\bibitem[Moffaert and Now{\'{e}}(2014)]{Moffaert2014}
Kristof~Van Moffaert and Ann Now{\'{e}}.
\newblock {Multi-Objective Reinforcement Learning using Sets of Pareto Dominating Policies}.
\newblock \emph{Journal of Machine Learning Research}, 15:\penalty0 3663--3692, 2014.

\bibitem[Ouyang et~al.(2022)Ouyang, Wu, Jiang, Almeida, Wainwright, Mishkin, Zhang Sandhini Agarwal Katarina Slama Alex Ray John Schulman Jacob Hilton Fraser Kelton Luke Miller Maddie Simens Amanda~Askell, Welinder Paul~Christiano, Leike, and Lowe]{Ouyang2022}
Long Ouyang, Jeff Wu, Xu~Jiang, Diogo Almeida, Carroll~L Wainwright, Pamela Mishkin, Chong Zhang Sandhini Agarwal Katarina Slama Alex Ray John Schulman Jacob Hilton Fraser Kelton Luke Miller Maddie Simens Amanda~Askell, Peter Welinder Paul~Christiano, Jan Leike, and Ryan Lowe.
\newblock {Training language models to follow instructions with human feedback}.
\newblock \emph{Advances in neural information processing systems}, 35:\penalty0 27730--27744, 2022.

\bibitem[Peschl et~al.(2022)Peschl, Zgonnikov, Oliehoek, and Siebert]{Peschl2021}
Markus Peschl, Arkady Zgonnikov, Frans~A. Oliehoek, and Luciano~C. Siebert.
\newblock {MORAL: Aligning AI with Human Norms through Multi-Objective Reinforced Active Learning}.
\newblock In \emph{AAMAS 2022: 21st International Conference on Autonomous Agents and Multiagent Systems (Virtual)}, pages 1038--1046. International Foundation for Autonomous Agents and Multiagent Systems, 12 2022.
\newblock URL \url{http://arxiv.org/abs/2201.00012}.

\bibitem[Rame et~al.(2023)Rame, Couairon, Dancette, Gaya, Shukor, Soulier, and Cord]{Rame2023}
Alexandre Rame, Guillaume Couairon, Corentin Dancette, Jean-Baptiste Gaya, Mustafa Shukor, Laure Soulier, and Matthieu Cord.
\newblock {Rewarded soups: towards Pareto-optimal alignment by interpolating weights fine-tuned on diverse rewards}.
\newblock In A~Oh, T~Naumann, A~Globerson, K~Saenko, M~Hardt, and S~Levine, editors, \emph{Advances in Neural Information Processing Systems}, volume~36, pages 71095--71134. Curran Associates, Inc., 2023.
\newblock URL \url{https://proceedings.neurips.cc/paper_files/paper/2023/file/e12a3b98b67e8395f639fde4c2b03168-Paper-Conference.pdf}.

\bibitem[Sorensen et~al.(2024{\natexlab{a}})Sorensen, Jiang, Hwang, Levine, Pyatkin, West, Dziri, Lu, Rao, Bhagavatula, Sap, Tasioulas, and Choi]{Sorensen2023}
Taylor Sorensen, Liwei Jiang, Jena Hwang, Sydney Levine, Valentina Pyatkin, Peter West, Nouha Dziri, Ximing Lu, Kavel Rao, Chandra Bhagavatula, Maarten Sap, John Tasioulas, and Yejin Choi.
\newblock {Value Kaleidoscope: Engaging AI with Pluralistic Human Values, Rights, and Duties}.
\newblock In \emph{Proceedings of the AAAI Conference on Artificial Intelligence}, pages 19937--19947, 9 2024{\natexlab{a}}.
\newblock \doi{10.1609/aaai.v38i18.29970}.
\newblock URL \url{http://arxiv.org/abs/2309.00779 http://dx.doi.org/10.1609/aaai.v38i18.29970}.

\bibitem[Sorensen et~al.(2024{\natexlab{b}})Sorensen, Moore, Fisher, Gordon, Mireshghallah, Rytting, Ye, Jiang, Lu, Dziri, Althoff, and Choi]{Sorensen2024}
Taylor Sorensen, Jared Moore, Jillian Fisher, Mitchell Gordon, Niloofar Mireshghallah, Christopher~Michael Rytting, Andre Ye, Liwei Jiang, Ximing Lu, Nouha Dziri, Tim Althoff, and Yejin Choi.
\newblock {Position: A Roadmap to Pluralistic Alignment}.
\newblock In \emph{Forty-first International Conference on Machine Learning}, 2024{\natexlab{b}}.

\bibitem[Sutton and Barto(2018)]{Sutton2018}
Richard~S Sutton and Andrew~G Barto.
\newblock \emph{{Reinforcement learning: an Introduction}}.
\newblock MIT press, 2018.

\bibitem[Taylor et~al.(2021)Taylor, Berrueta, and Murphey]{Taylor2021}
Annalisa~T. Taylor, Thomas~A. Berrueta, and Todd~D. Murphey.
\newblock {Active learning in robotics: A review of control principles}.
\newblock \emph{Mechatronics}, 77, 8 2021.
\newblock ISSN 09574158.
\newblock \doi{10.1016/j.mechatronics.2021.102576}.

\bibitem[Taylor et~al.(2020)Taylor, Yudkowsky, LaVictoire, and Critch]{Taylor2020}
Jessica Taylor, Eliezer Yudkowsky, Patrick LaVictoire, and Andrew Critch.
\newblock {Alignment for Advanced Machine Learning Systems}.
\newblock In \emph{Ethics of Artificial Intelligence}, pages 342--382. Oxford University Press, 9 2020.
\newblock \doi{10.1093/oso/9780190905033.003.0013}.
\newblock URL \url{https://academic.oup.com/book/33540/chapter/287906349}.

\bibitem[Vamplew et~al.(2017)Vamplew, Issabekov, Dazeley, Foale, Berry, Moore, and Creighton]{Vamplew2017b}
Peter Vamplew, Rustam Issabekov, Richard Dazeley, Cameron Foale, Adam Berry, Tim Moore, and Douglas Creighton.
\newblock {Steering approaches to Pareto-optimal multiobjective reinforcement learning}.
\newblock \emph{Neurocomputing}, 263:\penalty0 26--38, 11 2017.
\newblock ISSN 09252312.
\newblock \doi{10.1016/j.neucom.2016.08.152}.

\bibitem[Vamplew et~al.(2018)Vamplew, Dazeley, Foale, Firmin, and Mummery]{Vamplew2018}
Peter Vamplew, Richard Dazeley, Cameron Foale, Sally Firmin, and Jane Mummery.
\newblock {Human-aligned artificial intelligence is a multiobjective problem}.
\newblock \emph{Ethics and Information Technology}, 20\penalty0 (1), 2018.
\newblock ISSN 15728439.
\newblock \doi{10.1007/s10676-017-9440-6}.

\bibitem[Yang et~al.(2024)Yang, Pan, Luo, Qiu, Zhong, Yu, and Chen]{Yang2024}
Rui Yang, Xiaoman Pan, Feng Luo, Shuang Qiu, Han Zhong, Dong Yu, and Jianshu Chen.
\newblock {Rewards-in-Context: Multi-objective Alignment of Foundation Models with Dynamic Preference Adjustment}.
\newblock \emph{ArXiv preprint arXiv:2402.10207}, 2024.

\end{thebibliography}

}

\end{document}